\documentclass[11pt]{article}
\usepackage{fullpage}
\makeatletter
\long\def\@makecaption#1#2{
  \vskip 0.8ex
  \setbox\@tempboxa\hbox{\small {\bf #1:} #2}
  \parindent 1.5em  
  \dimen0=\hsize
  \advance\dimen0 by -3em
  \ifdim \wd\@tempboxa >\dimen0
  \hbox to \hsize{
    \parindent 0em
    \hfil 
    \parbox{\dimen0}{\def\baselinestretch{0.96}\small
      {\bf #1.} #2
    } 
    \hfil}
  \else \hbox to \hsize{\hfil \box\@tempboxa \hfil}
  \fi
}
\makeatother

\usepackage[pdftex]{graphicx}
\usepackage{amsmath,amssymb,amsthm}
\usepackage{diagbox}
\usepackage{multirow}
\usepackage{subcaption}
\usepackage[ruled,vlined]{algorithm2e}

\SetCommentSty{mycommfont}

\usepackage{amsfonts}
\usepackage{amsmath}
\usepackage{amsthm}
\usepackage{amssymb}
\usepackage{caption}
\usepackage{microtype}
\usepackage{graphicx}
\usepackage{subcaption}
\usepackage{booktabs} 
\usepackage[numbers]{natbib}
\usepackage{fancyhdr}
\usepackage{color}
\usepackage{forloop}

\usepackage[utf8]{inputenc} 
\usepackage[T1]{fontenc}    
\usepackage{hyperref}       
\usepackage{url}            
\usepackage{booktabs}       
\usepackage{amsfonts}       
\usepackage{nicefrac}       
\usepackage{microtype}      
\usepackage{xcolor}         
\usepackage{bm}
\usepackage{pifont,dsfont}
\usepackage{placeins}
\usepackage{enumitem} 
\usepackage{xspace}

\newcommand{\method}{MET\xspace}
\newcommand{\met}{\method\xspace}
\newcommand{\methodstd}{MET-S }

\newcommand{\mnist}{MNIST\xspace}
\newcommand{\cifar}{CIFAR-10\xspace}
\newcommand{\fmnist}{FMNIST\xspace}
\newcommand{\covtype}{CovType\xspace}
\newcommand{\adult}{Adult-Income\xspace}
\newcommand{\R}{\mathbb{R}}
\newcommand{\E}{\mathbb{E}}
\newcommand{\cD}{\mathcal{D}}
\newcommand{\f}{f_{\theta}}
\newcommand{\h}{h_{\phi}}

\newcommand{\abs}[1]{\left|#1\right|}
\newcommand{\norm}[1]{\left\|#1\right\|}

\usepackage{hyperref}

\begin{document}

\abovedisplayskip=8pt plus0pt minus3pt
\belowdisplayskip=8pt plus0pt minus3pt

\begin{center}
  {\LARGE \textbf{MET: Masked Encoding for Tabular Data}} \\
  \vspace{.5cm}
  {\Large Kushal Majmundar$^{\star}$ ~ Sachin Goyal$^{\dagger}$ ~ Praneeth Netrapalli$^{\star}$ ~ Prateek Jain$^{\star}$} \\
  \vspace{.4cm}
  {\large $^{\star}$Google Research, India, \large $^{\dagger}$Carnegie Mellon University} \\
  \vspace{.4cm}
  \texttt{\{majak,pnetrapalli,prajain\}@google.com, sachingo@andrew.cmu.edu}
\end{center}

\begin{abstract}
We consider the task of self-supervised representation learning (SSL) for tabular data -- tabular-SSL. Typical contrastive learning based  SSL methods require instance-wise data augmentations which are difficult to design for unstructured tabular data. Existing tabular-SSL methods design such augmentations in a relatively ad-hoc fashion and can fail to capture the underlying data manifold. Instead of augmentations based approaches for tabular-SSL, we propose a new reconstruction based method, called \emph{Masked Encoding for Tabular Data (\method)}, that does not require augmentations. \method is based on the popular MAE approach for vision-SSL \cite{mae} and uses two key ideas: (i) since each coordinate in a tabular dataset has a distinct meaning, we need to use separate representations for all coordinates, and (ii) using an adversarial reconstruction loss in addition to the standard one.
Empirical results on five diverse tabular datasets show that \method achieves a new state of the art (SOTA) on all of these datasets and improves up to 9\% over current SOTA methods. We shed more light on the working of \method via experiments on carefully designed simple datasets.
\end{abstract}
\section{Introduction}\label{sec:introduction}

Recently, self-supervised pre-training (SSL) followed by supervised fine-tuning has emerged as the state of the art approach for semi-supervised learning in domains such as natural language processing (NLP)~\cite{bert}, computer vision~\cite{simclr} and speech/audio processing~\cite{wav2vec2}.
Given that there is an extensive amount of raw, unlabeled data in various settings such as healthcare, finance, marketing, etc., most of which exist in tabular form, extending SSL to tabular data is an important direction of research.

Broadly speaking, there are two dominant approaches to SSL: (i) reconstruction of masked inputs, and (ii) invariance to certain augmentations/transformations, also known as \emph{contrastive learning}. Several prior works~\cite{dacl,ucar2021subtab} have adopted the second approach of contrastive learning for designing SSL methods for tabular data (tabular-SSL). The underlying structure and semantics of specific domains such as images remain somewhat static, irrespective of the dataset. So, one can design generalizable domain specific augmentations like cropping, rotating, resizing etc. However, tabular data does not have such fixed input vocabulary space (such as pixels in images) and semantic structure, and thus lacks generalizable augmentations across different datasets. 
Consequently, there are only a limited number of augmentations that have been proposed for the tabular setting such as mix-up, adding random (gaussian) noise and selecting subsets of features~\cite{dacl,ucar2021subtab}. While reconstruction based SSL methods have also been proposed earlier for tabular data~\cite{yoon2020vime}, their performance is suboptimal compared to the contrastive based approaches~\cite{dacl,ucar2021subtab}.

In this paper, we build upon recent advances in reconstruction based SSL methods in computer vision~\cite{mae} to design \method -- a \emph{purely reconstruction based} approach for tabular-SSL. \method achieves a new state of the art (SOTA) performance on several tabular datasets. In particular, \method gives an average improvement of $3.2\%$ accuracy (averaged over five standard datasets) over the previous state-of-the-art approaches. 

There are two key ideas behind our approach. 
Similar to the reconstruction based approaches for SSL of images~\cite{mae} and text~\cite{bert}, we use a transformer architecture to efficiently learn the relationships between different coordinates. However, \emph{unlike}~\cite{mae,bert}, using an average pooling or a special output token for the representations leads to a large drop in accuracy on tabular datasets in general. So, our first idea is to instead \emph{concatenate embeddings of all tokens for the final finetuning step}.  Surprisingly, even though the finetuning step trains a large model, it generalizes very well. {We conjecture, and empirically demonstrate on a simple toy-dataset, that the highly separable representations alleviates the risk of overfitting during the fine-tuning phase.} Second, our proposed method \method searches for adversarial perturbations in the input space, by performing gradient ascent over reconstruction loss, which we add to the input before passing to the auto-encoder. We observe that adversarial training along with the masked reconstruction loss gives superior representations, which have better downstream classification accuracy.

We conduct thorough experiments to demonstrate the effectiveness of \method in the standard tabular-SSL setting \cite{dacl}. On standard tabular-SSL benchmarks like permuted \mnist, permuted \fmnist, permuted \cifar, as well as on other popular tabular datasets such as \adult and \covtype, we show that \method can be up to 10\% more accurate than the SOTA tabular-SSL methods like DACL \cite{dacl}, SubTab \cite{ucar2021subtab}, VIME  \cite{yoon2020vime}. Furthermore, in some cases, \method trained with about 20\% of the labelled train-set can be as effective as a standard supervised learning methods trained with all the labelled points in the train-set. Finally, we demonstrate that \method indeed learns non-trivial powerful representations which are significantly more powerful than the random kitchen-sink features which are known to approximate the universal RBF kernel \cite{kitchensink}.   

To summarize, in this paper, we design a novel algorithm for tabular-SSL. The algorithm is based on three key insights (i) masking is a natural technique to bottleneck the information in tabular data, (ii) utilizing the embeddings for all coordinates without average pooling, (ii) an adversarial reconstruction loss. Experiments on {five different datasets} demonstrate the efficacy of our algorithm in practice, and that it significantly improves over SOTA methods.


\setlength{\textfloatsep}{5pt}
\begin{figure}
  \centering
  \includegraphics[width=\linewidth]{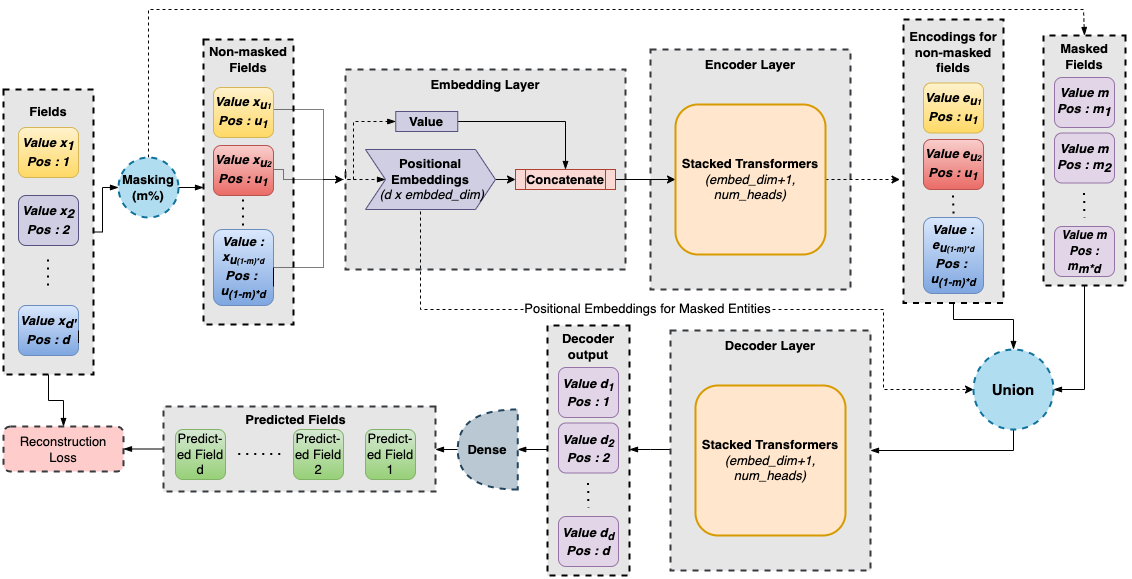}\vspace*{-5pt}
  \caption{MET Framework for tabular-SSL. Given an input, we mask out a fraction of co-ordinates (features). The masked input is then concatenated with its learnable positional encodings and fed to the transformer based encoder as input. The obtained encoder output (learnt representations) are then passed through the decoder along with the mask token. Recontruction loss is then optimized end-to-end.}
  \label{fig:mate}
\end{figure}
\section{Related Work}

\paragraph{Self-Supervised Learning : } Self supervised learning (SSL) has shown promising results not only in the regimes where the labelled training data is scarce but has also shown great empirical success in training large scale models across various domains like Natural Language Processing and Computer Vision. SSL can be broadly classified into two categories : Pretext task based approaches and contrastive learning based approaches. Pretext based SSL approaches solve a "pretext" task like reconstruction from a masked or a noisy input, in order to learn the underlying distribution of the unlabelled data. Wav2Vec\cite{wav2vec2} efficiently trains a large scale speech-to-text model using masking based reconstruction for learning good speech representations. Similarly, \cite{mask_speech, hubert} have used masking for learning speech representations. Masked language modelling has been extensively studied in literature \cite{bert, radford}, and has shown quite promising results. Motivated by the success of masking based approaches in NLP and speech, a recent paper \cite{mae} proposed a masked input reconstruction approach for visual representation learning. Prior to this, \cite{mst} also proposed masking although in feature space. In this paper, motivated by the constraints in tabular setting along with the success of masking based approaches, we build a purely reconstruction based approach for tabular-SSL. 

A concurrent line of work in SSL learns representations using instance level separation tasks as discussed in \cite{simclr, simsiam, barlow, moco}. Some other pretext tasks like solving jigsaw puzzles have also been proposed in \cite{jigsaw}. However, all these approaches require domain-specific knowledge to create positive-negative sample pairs. Some recent advances towards a domain agnostic approach have also been proposed like \cite{dacl}. We compare our proposed algorithm \method against such approaches in Section \ref{sec:experiments}. 
 
\paragraph{Adversarial Self supervised learning} While adversarial SSL has been explored in the context of contrastive learning~\cite{kim2020adversarial}, it seems to be less explored for reconstruction based SSL. Our method \method proposes a novel framework where we try to find adversarial points in the input manifold which have a high reconstruction loss. \cite{chen2020adversarial} have also proposed adversarial learning, although to learn robust pre-trained models. \method instead explores the use of adversarial search over input manifold to learn better separable representations for higher accuracy on downstream classification.
\cite{adv_mask} proposes to find an adversarial mask which maximizes the distance between the representations of input and its masked (adversarial) counterpart.

\paragraph{Self Supervised Learning for Tabular Data} Reconstruction based SSL has been previously explored in SubTab \cite{ucar2021subtab}, which treats it as a multi-view representation learning problem. They try to learn representations for multiple croppings of the input data and at inference time aggregate the representations of the croppings (multiple-views). Note that \method performs random masking over the input space only at the training time, at inference the representation is given by passing all the co-ordinates through the encoder and hence does not consider the problem as a multi-view representation learning. \cite{yoon2020vime} uses a combination of predicting the masked tokens and reconstruction. Both \cite{yoon2020vime, ucar2021subtab} use gaussian noise addition to the input to prevent the auto-encoder from learning an identity mapping. \method efficiently searches for noise using adversarial search to learn better representations.

\section{Preliminaries}
In this section, we formalize the general task of learning self-supervised representations and introduce all the notations and assumptions applicable.

\textbf{Notation:} We use $x_i$ to denote an example and $x_i^j$ to denote the $j^{\textrm{th}}$ coordinate of $x_i$. For a set $S$ of coordinates, $x_i^S$ denotes the restriction of $x_i$ to the coordinates represented by $S$.

\textbf{Task:} Consider access to a corpus of unlabelled dataset given by $\mathcal{D}_u=\{x_i\}_{i=1}^{N_u}$ where each datapoint $x_i\in\mathbb{R}^d$. Further, every co-ordinate in $x_i$ i.e. $x_i^j\in\mathbb{R}$ can be either a categorical or a non-categorical value, without being explicitly specified. The general goal of self-supervised learning is to learn a parameterized mapping $f_{\theta}: \mathbb{R}^d \rightarrow \mathbb{R}^m$ between the input $x_i$ and its representation $f_{\theta}(x_i)\in\mathbb{R}^m$, such that the representations are well suited for a downstream task as described next.

\textbf{Evaluation of learned representations:} In this paper, we evaluate the quality of learned representations through accuracy on a downstream classification task. More concretely, we have access to a small set of labelled training dataset $\mathcal{D}_\text{train}=\{x_i,y_i\}_{i=1}^{N_\text{train}}$ where $y_i\in \mathbb{R}^k$ and each $(x_i,y_i)$ is drawn independently and identically (i.i.d.) from some underlying distribution $\mathcal{D}$ on $\R^d \times \R^k$.
The task is to learn a classifier $h_{\phi}: \mathbb{R}^d \rightarrow \mathbb{R}^k$ which minimizes $ \E_{(x,y)\sim\cD}\left[\ell(h_\phi(x),y)\right]$, where $\ell$ is a loss function such as $0-1$ loss or cross entropy loss etc.
Given the learned representations $f_\theta$, we train a shallow classifier $g_{\mu}: \mathbb{R}^m \rightarrow \mathbb{R}^k$ (we use a $2$-hidden layer MLP in our default setting) and use the resulting accuracy to evaluate the quality of $f_\theta$.




\section{Method}
\label{sec:method}
As described in the previous section, our goal is to learn a parameterized mapping $f_{\theta}$. We take a denoising auto-encoder approach for this~\cite{vincent2008extracting}. In this approach, 
we have an encoder represented by $\f$ and a decoder represented by $h_{\phi}$. The task of the encoder is to take a noisy version of input example $x_i$, e.g., where some coordinates of $x_i$ are masked, and reconstruct the entire example $x_i$.
More formally, if we choose $S_i \subseteq [d]$ coordinates to be masked in $x_i$, then this approach can be written as minimization of the following function:
\begin{equation}
    \mathcal{L}_\text{rec}(\theta,\phi)=\sum_{i=1}^{N_u}\|x_i - \h(\f(S_i, x_i^{S_i}))\|_2^2. \label{eqn:recloss}
\end{equation}
While the high level approach of denoising autoencoder was proposed in the seminal paper~\cite{vincent2008extracting}, instantiating this approach for various domains has required several domain specific insights including architectures of $\h$ and $\f$, which coordinates to mask etc. (~\cite{bert,van2018representation,mae}). For downstream evaluation task, we discard the decoder $\h$ and only use the representations computed by $\f$. We now describe the details of the architectures of the encoder and decoder as well as the masking strategy.

\begin{algorithm}[t]
\DontPrintSemicolon

\SetKwInOut{Input}{Input}
\SetKwInOut{Arch}{Architecture}

\Input{Unlabelled data $\mathcal{D}_u=\{x_i\}_{i=1}^{N_u}$, Set of masked co-ordinates $\{S_i\}_{i=1}^{N_u}$, Encoder $\f$, Decoder $\h$, projection radius $\epsilon$, weight of adversarial loss $\lambda$}
\For{$iteration = 0, 1,$ ... $N-1$}
{
    \For{$x_i \in \mathcal{D}_u$}
    {
        $\hat{x}_i= \h(\f(S_i, x_i^{S_i}))$ \tcc*{Try to reconstruct from masked input.}
        $\mathcal{L}_\text{rec}^\textrm{std} = \|x_i - \hat{x}_i\|_2^2$ \tcc*{Standard reconstruction loss.}
        $h\sim \mathcal{N}(0, \mathcal{I}_d)/\sqrt{d}$ \tcc*{Initialize adversarial perturbation.}
        \For{steps in $1, 2, \hdots \text{adv\_steps}$}
            {
            \tcc{Find adversarial perturbation $h$ to maximize reconstruction loss using gradient ascent.}
            $\hat{x}_i= \h(\f(S_i, (x_i+h)^{S_i}))$\;
            $\mathcal{L}_\text{rec}(h) = \|x_i - \hat{x}_i\|_2^2$\;
            $h = h + \eta_{1} \frac{\nabla_h \mathcal{L}_\text{rec}}{\|\nabla_h \mathcal{L}_\text{rec}\|}$\;
            $h = \frac{h}{\|h\|} \alpha$ where $\alpha = \|h\|\cdot\mathds{1}[\|h\|<\epsilon] + \epsilon\cdot\mathds{1}[\|h\|\geq \epsilon]$
            }
        $\hat{x}_i= \h(\f(S_i, (x_i+h)^{S_i}))$\;
        $\mathcal{L}_\text{rec}^\text{adv} = \|x_i - \hat{x}_i\|_2^2$ \tcc*{Adversarial reconstruction loss.}
        $\mathcal{L}_\text{total} = \mathcal{L}_\text{rec}^\textrm{std} + \lambda \cdot \mathcal{L}_\text{rec}^\text{adv}$ \tcc*{Final loss is a sum of standard and adversarial reconstruction losses.}
        $(\theta, \phi) = (\theta,\phi) - \eta_{2} (\nabla_{\theta} \mathcal{L}_\text{total}, \nabla_{\phi} \mathcal{L}_\text{total})$ \tcc*{Gradient descent on $\theta$ and $\phi$.}
    }
}
\caption{MET : Masked Encoding Tabular data}
\label{algorithm:meta}
\end{algorithm}
\textbf{Encoder-Decoder architecture:} Motivated by the success of transformers~\cite{vaswani2017attention} across various domains such as natural language processing (NLP) and computer vision, we use a transformer as our backbone for both the encoder as well as the decoder. Recall that $\f$ and $\h$ denote the encoder and decoder respectively. The input to a transformer is given as a set of tokens, each represented by an embedding. In the current context, given an example $x_i$, each coordinate $j\in [d]$ is masked with some probability $p$. Recall that $S_i$ denotes the set of masked coordinates for $x_i$. The input to the transformer consists of $\abs{[d]\setminus S_i}$ tokens each corresponding to one coordinate $j\in [d]\setminus S_i$. Note that no token is passed corresponding to the masked coordinates. The input embedding corresponding to the $j^{\textrm{th}}$ token consists of a concatenation of $pe_j \in \R^e$, a learnable encoding corresponding to the $j^{\textrm{th}}$ coordinate and $x_i^j$, the value of $j^{\textrm{th}}$ coordinate in $x_i$. So, the total dimension of the embedding is $e+1$. The encoder takes these input tokens through several layers of multi-headed attention and feed-forward layer of dimension $fw$ and outputs an $e+1$ dimensional representation $w_i^j$ for each token $j\in [d]\setminus S_i$. We now pass these to the decoder $\h$, along with representations for masked coordinates -- again concatenation of $pe_j$ with a learnable mask parameter $u \in \R$ for $j\in S_i$.
\textbf{Adversarial loss}: In the context of supervised learning, several papers have demonstrated that adversarial training can yield more robust features~\cite{tsipras2018robustness} that are better for transfer learning~\cite{salman2020adversarially}. While an adversarial loss function has been observed to encourage learning of robust features in contrastive SSL~\cite{chen2020adversarial,kim2020adversarial}, to the best of our knowledge, it does not seem to have been explored in the context of masked autoencoders. In this work, we demonstrate that an adversarial version of the reconstruction loss works better than standard reconstruction loss. More concretely, the adversarial reconstruction loss is given by:
\begin{align}
    \mathcal{L}_\text{rec}^\textrm{adv}(\theta,\phi)=\sum_{i=1}^{N_u} \max_{\delta : \norm{\delta} \leq \epsilon}\|x_i - \h(\f(S_i, x_i^{S_i}+\delta))\|_2^2. \label{eqn:adv-rec}
\end{align}
In this paper, we constrain the adversarial noise $\delta$ in an $\epsilon$ radius $L2$ norm ball around the input data point $x_i$, where $\epsilon$ is chosen from a grid-search in $ \{2,4,6,10,12,14\}$. The overall algorithm for \method, which  minimizes~\eqref{eqn:adv-rec} and ~\eqref{eqn:recloss} is given in Algorithm~\ref{algorithm:meta}. For consistency of notation, we present a non-batch version of the algorithm.

\begin{figure}
    \centering
  \begin{subfigure}{0.25\textwidth}
  \centering
    \includegraphics[height=2.7cm]{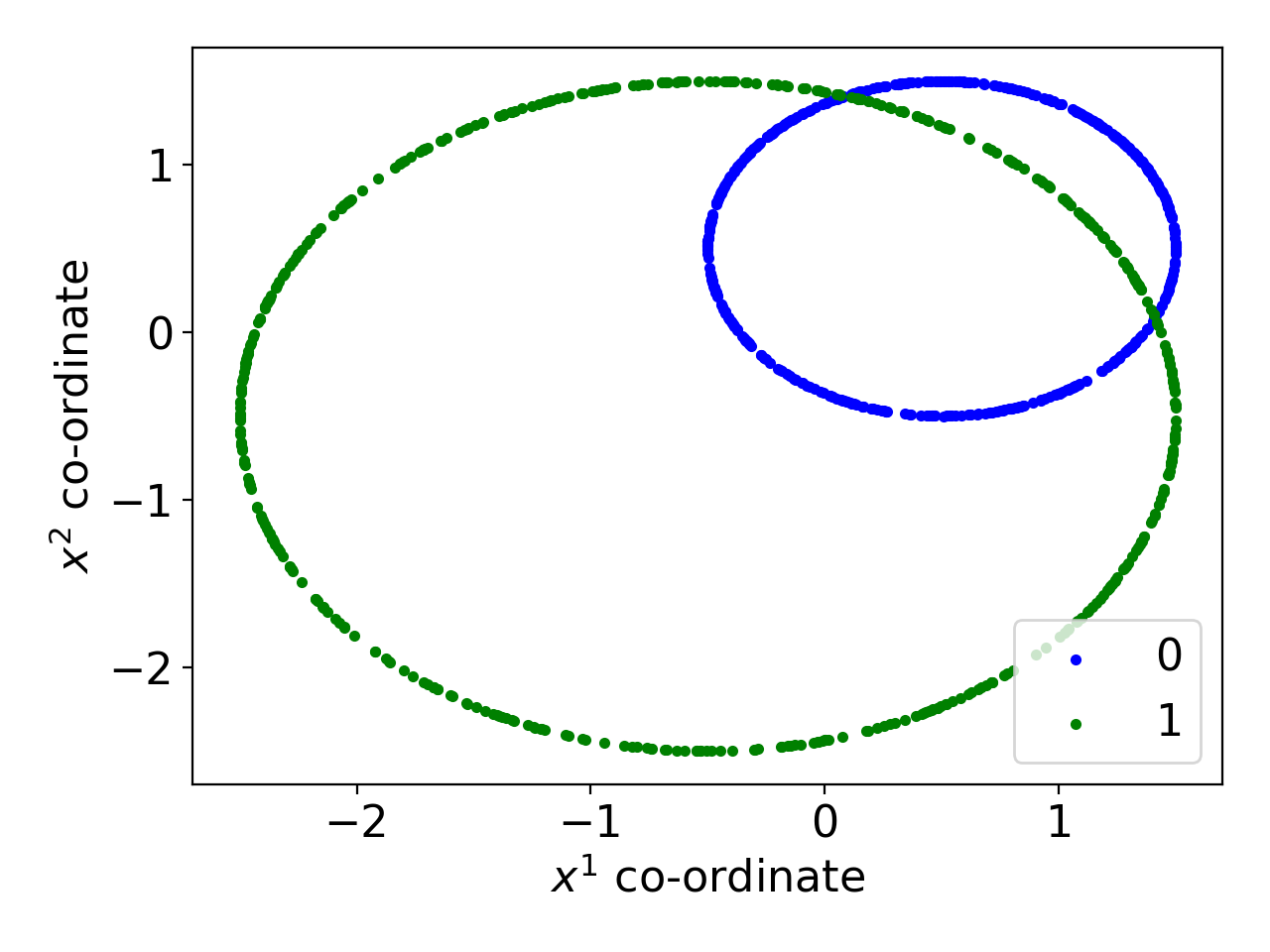}
    \caption{}\label{fig:toy_data}
  \end{subfigure}%
  \hspace*{\fill}
  \begin{subfigure}{0.25\textwidth}
  \centering
    \includegraphics[height=2.7cm]{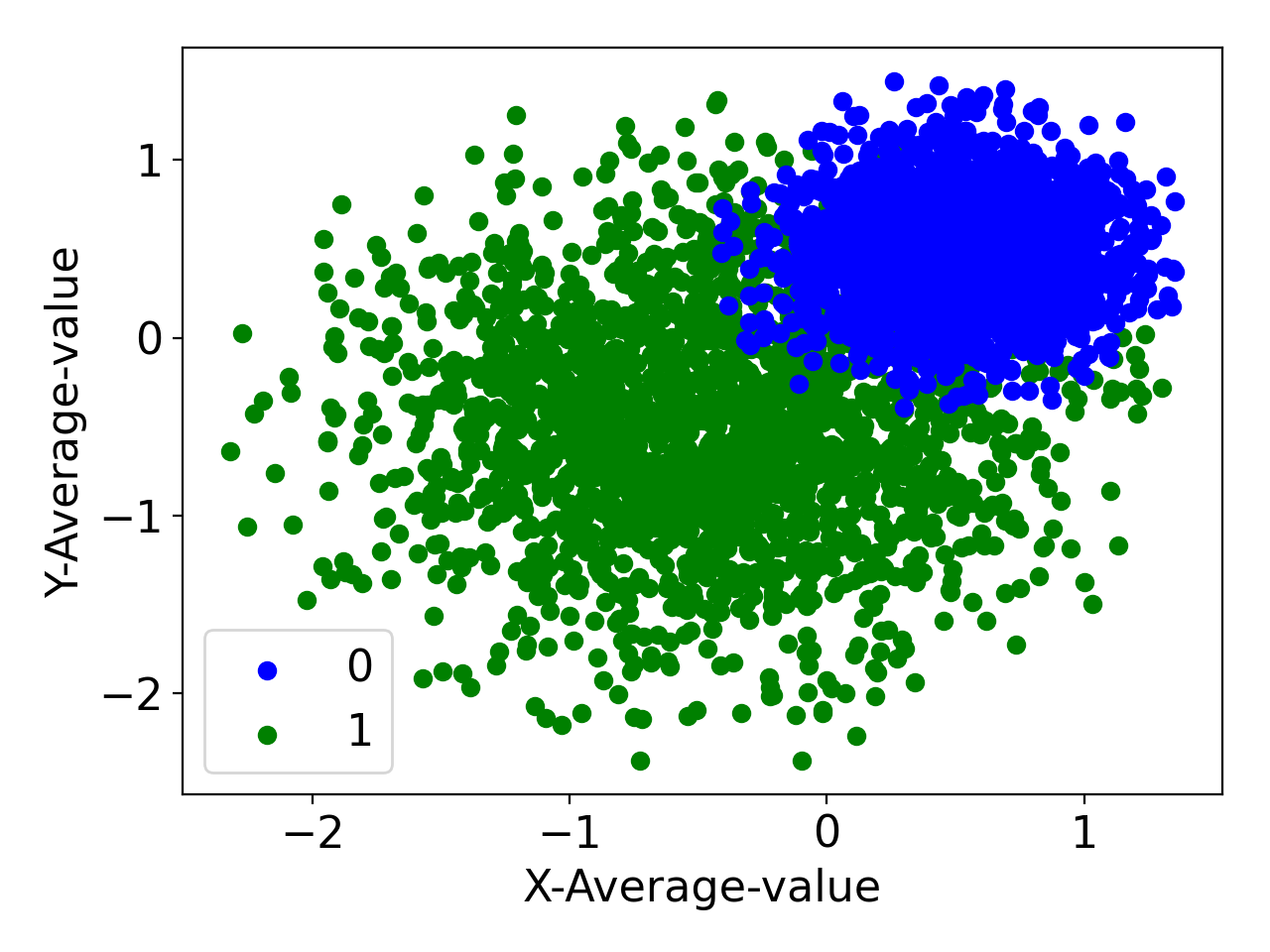}
    \caption{}\label{fig:synthetic_raw}
  \end{subfigure}%
  \hspace*{\fill}
  \begin{subfigure}{0.25\textwidth}
  \centering
    \includegraphics[height=2.7cm]{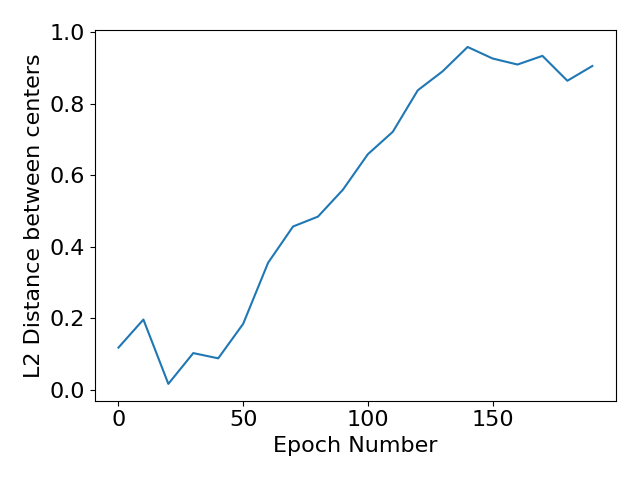}
    \caption{}\label{fig:toy_norm_dist}
  \end{subfigure}%
  \hspace*{\fill}
  \begin{subfigure}{0.25\textwidth}
  \centering
    \includegraphics[height=2.7cm]{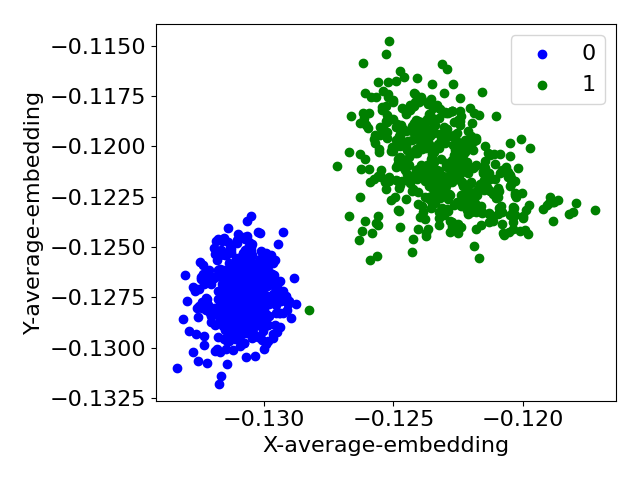}
    \caption{}\label{fig:toy_epoch_100}
  \end{subfigure}\vspace*{-5pt}
\caption{ (a) We analyze the representations learnt by \method on a 10-dimensional binary classification toy dataset, where 10-dimensional points are sampled by concatenating five points sampled i.i.d. from the respective circles. (b) 2D projection of the source data. (c): Mean distance between the learnt representations for the two classes as the SSL using \method proceeds. (d) 2D projection of the representations learnt by \method.}
\end{figure}

\subsection{Analysis on Toy Dataset}
Before moving to the experiments on standard tabular dataset benchmarks, we first work on a 10-dimensional toy dataset to visualize the kind of representations learnt by \method. We generate a binary classification dataset from two overlapping circles as shown in Figure\ref{fig:toy_data}. We generate 5000 samples for each class, where every 10-dimensional sample is a concatenation of five points sampled i.i.d from the respective circle.

Figure~\ref{fig:toy_norm_dist} shows the mean distance between the representations from the two classes as the self-supervised training using \method evolves. Notice that the separation between the learnt representations increases as the training progresses.

Further, for better visualization we project the generated data to 2D space, where the x and y axis projections for each 10-dimensional sample is obtained by taking average over alternate co-ordinates (since the data was generated by concatenating the x and y co-ordinates of five points sampled i.i.d from the circle). 
Figure~\ref{fig:synthetic_raw} and Figures~\ref{fig:toy_epoch_100} show the projections of the source data on the learnt representations on the 2D space. Observe that \method is able to learn representations which are much more easily separable compared to the source data.
\setlength{\textfloatsep}{5pt}
\section{Experiments}\label{sec:experiments}
Here, we empirically evaluate our proposed approach \method against SOTA tabular-SSL methods, along with other natural baselines as described in Section \ref{sec:baseline}. We experiment with  common tabular dataset benchmarks like the permuted MNIST, permuted FashionMNIST and permuted CIFAR-10. Further, we work with two other common tabular datasets from the UCI machine learning repository \cite{uci} : Forest CovType and Adult Income which are described in Section \ref{sec:data_des}.


\subsection{Implementation Details}
\label{sec:down_acc}
As discussed in Section \ref{sec:method}, the output of the encoder is used as the learnt representations for the downstream tasks like classification. Note that while training the downstream classifier, we do not mask the input i.e. concatenation of encoder output for all the co-ordinates of the input tabular data is used as the learnt representation. Using the labelled dataset, we train a MLP over the learnt representations for the downstream task, keeping the encoder fixed.

We use transformers \cite{att_all_u_need} as the backbone for both the encoder and the decoder. Embedding dimension for the encoder and the decoder is chosen from a gridsearch in $\{64,100,128\}$, feedforward dimension from  $\{64,100,128\}$, encoder and decoder depth from $\{1,3,6\}$ and the number of heads from $\{1,2,3\}$. The weight for adversarial reconstruction loss $\lambda$ in \method (see Algorithm~\ref{algorithm:meta}) is chosen to be $1$. All the remaining parameters are tuned using validation error on downstream task.
All the experiments have been performed on a cluster of Tesla P100 GPUs.

\subsection{Baselines and Existing Methods}\label{sec:baseline}
We compare \method with the following baselines:
\begin{itemize}[leftmargin=*]
    \item VIME \cite{yoon2020vime} : A SSL approach for tabular dataset, which uses a combination of masked token prediction and reconstruction loss.
    \item SubTab \cite{ucar2021subtab} : SubTab views SSL as a multi-view representation learning problem, where representations from multiple croppings are aggregated at test time. 
    
    \item DACL \cite{dacl} : A domain agnostic contrastive learning baseline which uses mixup noise as an augmentation. We specifically use DACL+ which uses geometric mean based mixup noise.
    
    \item MLP : We also compare against this natural baseline, wherein we train a MLP over the raw tabular data (and not the learnt representations) with the available labelled samples.
    \item Random Forest (RF): Following standard prescription for RF, we train a random forest with $100$ decision trees over the raw tabular data (and not the learnt representations) with the available labelled samples, until all leaf nodes have only one co-ordinate affecting it's decision criterion. 
    \item Random Featurization : To check the effectiveness of the learnt representations, we compare against fine-tuning an MLP over the representations from a random encoder i.e. a randomly initialized and fixed transformer. This is denoted by \method-R.
    \item Random Gaussian Featurization (RF-G) : Here, we compute  standard random kitchen sink \cite{kitchensink} style features, i.e., $\phi(x)=R x$ is the embedding of point $x\in \mathbb{R}^d$. Random features are known to be asymptotically an accurate approximation of the RBF kernel, which in turn is known to be a highly accurate and in fact, a ``universal" classifier for tabular data.  Note that we fix embedding dimension of RF-G to be same as that of \method. 

\end{itemize}

\subsection{Datasets}
\label{sec:data_des}
{\bf MNIST}\footnote{The data is normalized and shuffled for all datasets.}: The MNIST dataset of handwritten digits consists of 28x28 dimensional images, which are then flattened to get 784 coordinates in tabular form. The classification task consists of ten classes, one for each digit. A split of 60,000 entries as the train set and 10,000 entries as the test set is used as per the split for the original dataset.\\
{\bf FMNIST}: Fashion-MNIST(\textit{FMNIST}) is a dataset of Zalando's article images consisting of 28x28 dimensional images and is proposed as a more challenging replacement dataset for the MNIST dataset. The data is flattened to get 784 coordinates in tabular form and has ten classes. A split of 60,000 entries as the train set and 10,000 entries as the test set is used as per the split for the original dataset.\\
{\bf CIFAR-10}: 
The CIFAR-10 dataset contains 60,000 color images each of size 32x32 belonging to ten different classes with 6,000 images of each class. We flatten it to get 3072 coordinates in tabular form and use a split of 50,000 entries as the train set and 10,000 entries as the test set is used as per the split for the original dataset.\\
{\bf CoverType}: Forest CoverType(\textit{CoverType}) is a UCI dataset where the task is to predict forest cover type only from cartographic variables of a 30x30 meter cell, as determined from US Forest Service Region 2's resource information system. The data is not scaled and contains binary columns of data for qualitative independent variables: wilderness areas and soil types. It is a 7 class classification problem and consists of 54 features out of which one-hot vectors of wilderness area and soil type make up for 44 features. We replace them as two features by taking an argmax over the one-hot vectors and it reduces to 12 features in a tabular form. This makes the problem harder since the categorical features are now represented as integers instead of one-hot representations\footnote{Consequently, our accuracy numbers are not directly comparable to standard results on this dataset.}.
A split of 11,340 entries as train set and 565,892 entries as test set is used as per the split for the original dataset.\\
{\bf Income}: Adult Income(\textit{Income}) is a UCI dataset where the prediction task is to determine whether a person makes over \$50K a year based on census data. It consists of a mix of six continuous and eight categorical fields. Similar to CoverType dataset, we use integers instead of one hot representation for the categorical features and get 14 features in a tabular form . A split of 30,162 entries as train set and 15,060 as test set is used as per the split for the original dataset.


\subsection{Downstream Classification }
In this section, we compare \method against various baselines as mentioned in Section \ref{sec:baseline}. We compare the downstream classification accuracy of various representations. Specifically, we train an MLP over the learnt representations using all the available labeled dataset samples; see Section \ref{sec:down_acc}. 

\autoref{tab:classification} compares accuracy of \met with downstream classification against tabular-SSL methods and supervised learning baselines.  Both \method (adversarial noise + masking) and \methodstd (only masking) outperform the baselines across all the datasets. For example, on the permuted Fashion MNIST dataset, \method achieves an accuracy of $91.36\%$, outperforming all the other tabular SSL baselines like SubTab ($87.59\%$). Similarly, on CovType, \method is about $10\%$ more accurate DACL+, and in fact about $34\%$ more accurate than SubTab, perhaps due to lack of semantics in neighbouring columns in the dataset. Overall, we observe that \method gives an average improvement of $3.2\%$ accuracy compared to the nearest competitive baseline, establishing \method as a new state-of-the-art approach for self supervised learning on tabular data. 

Here, we would like to make two key points.\\ a) Note that using the same embedding dimensions, \met is able to give up to 18\% more accurate classifier than RF-G embeddings. This is interesting because RF-G embeddings are also able to capture non-linear features in the data; in fact, it can approximate RBF kernel itself. But, due to self-supervised training with the entire unlabeled data, \met can capture the data manifold more accurately, while RF-G are completely {\em independent} of the data distribution. This indicates the importance of further investigation of data-distribution based (random) embedding methods. \\
b) 
Here, we use a modified and much harder version of CovType, where the key categorical features like soil type are represented by their category index, instead of one-hot vectors. This immediately imposes an ordering on categories which is incorrect, and hence the representation learning method has to somehow learn to embed such coordinates in something similar to one-hot vectors, which in absence of additional domain information is challenging. Naturally, methods like MLP struggle on this dataset, but \met is able to get a reasonable accuracy which is 11\% higher than MLP. 

\begin{figure}
    \begin{minipage}{0.24\linewidth}
        \centering
        \includegraphics[width=\linewidth]{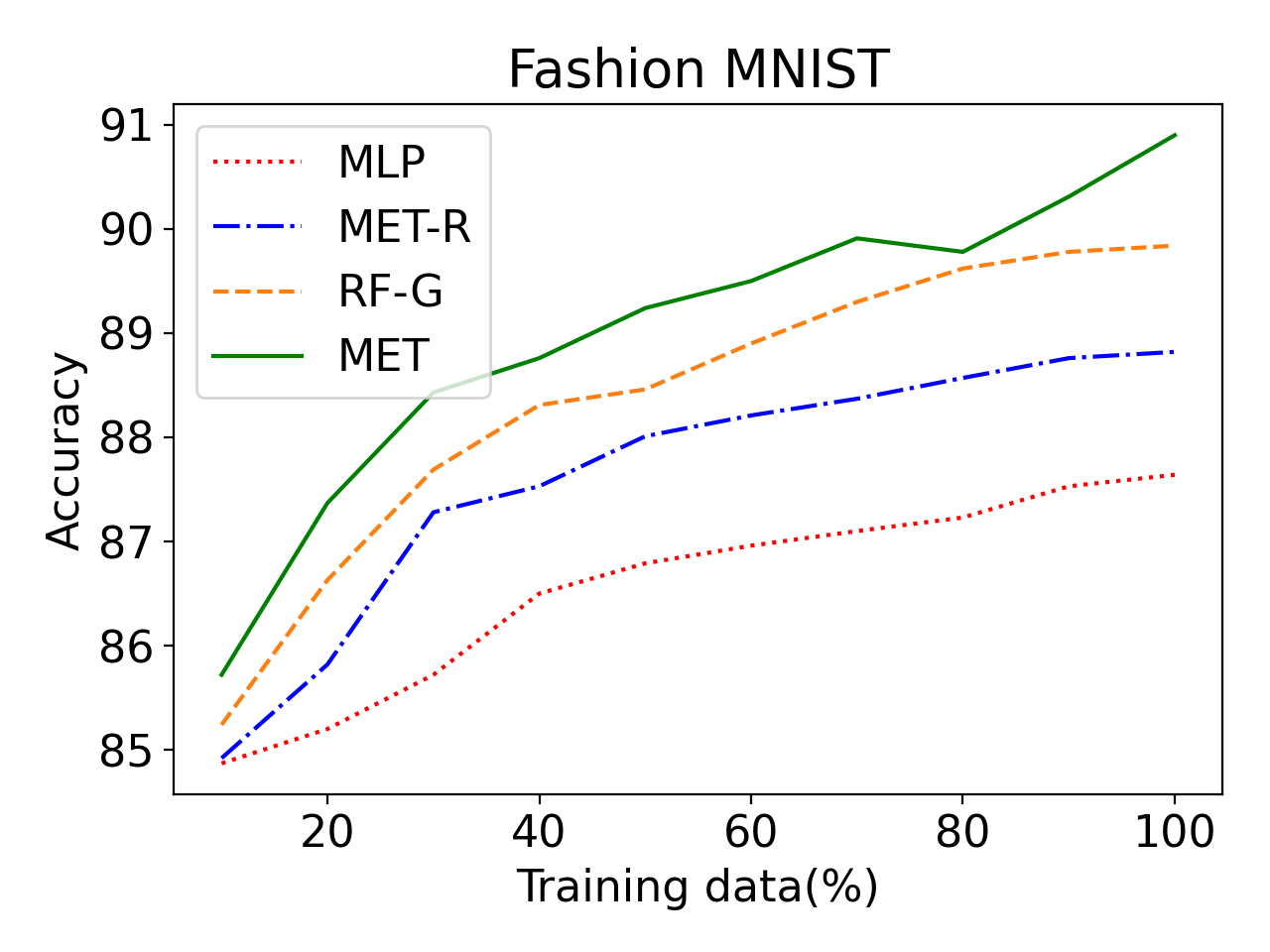}
        \end{minipage}%
        \begin{minipage}{0.24\linewidth}
        \centering
        \includegraphics[width=\linewidth]{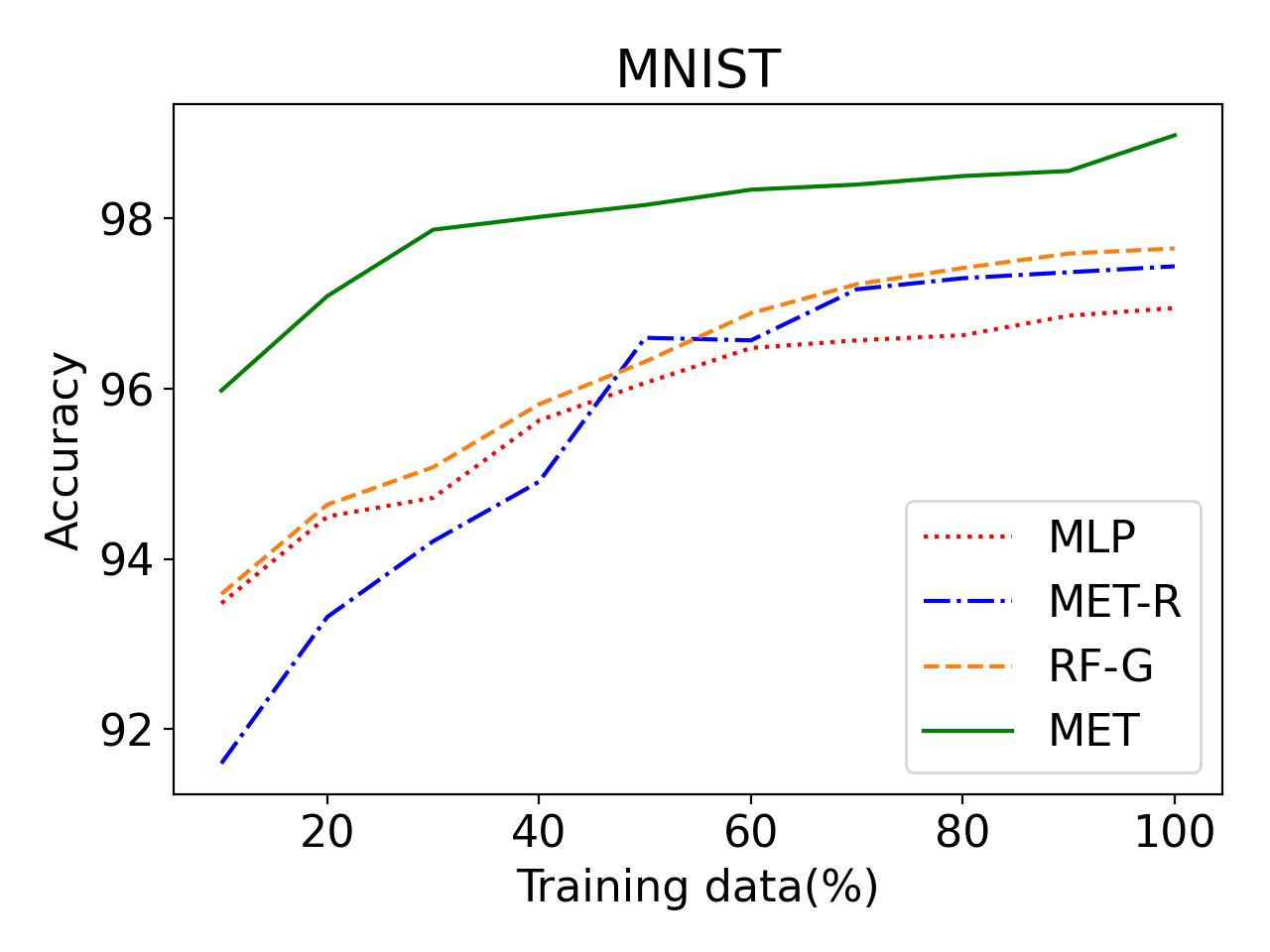}
    \end{minipage}
    \begin{minipage}{0.24\linewidth}
        \centering
        \includegraphics[width=\linewidth]{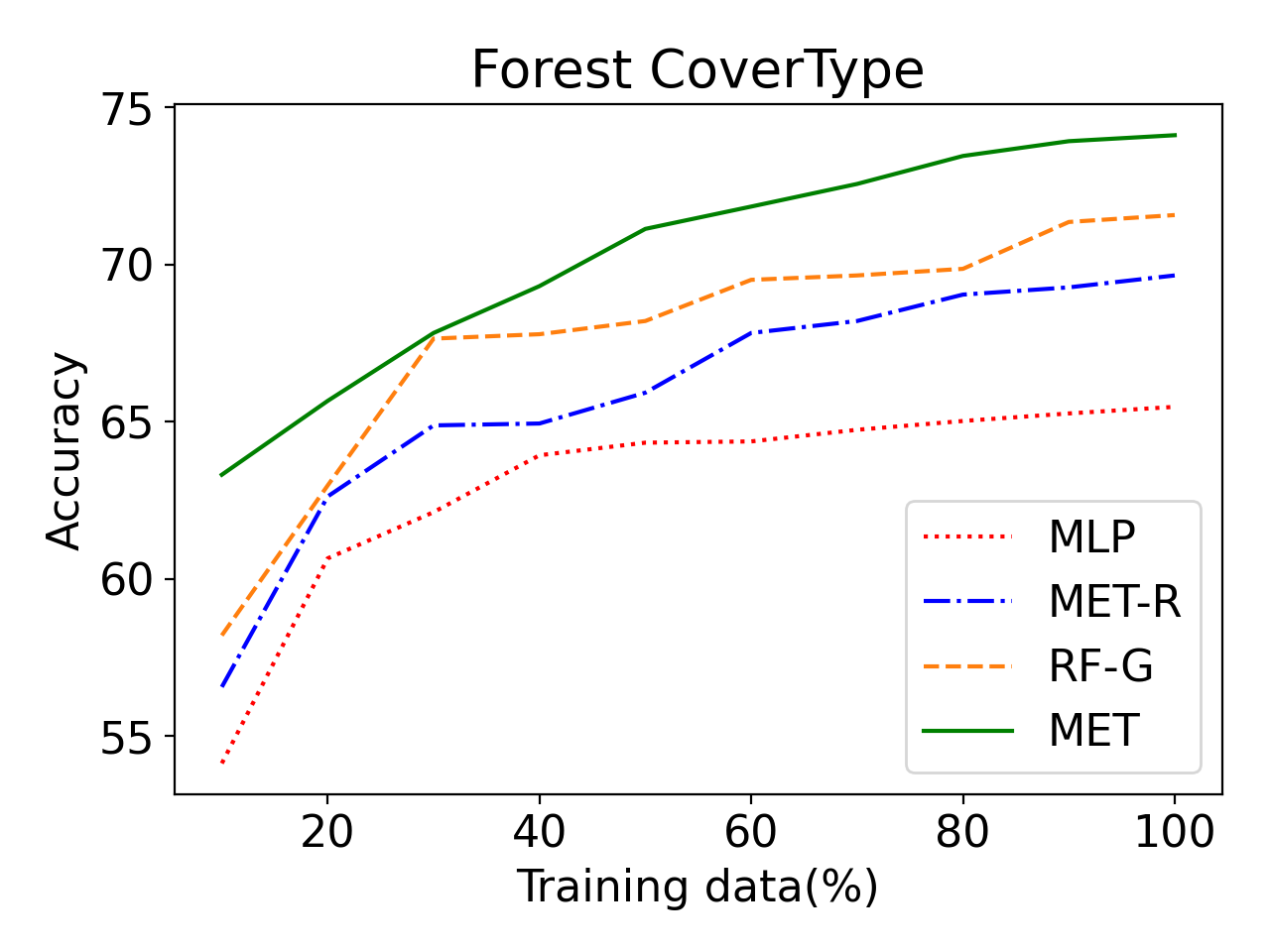}
        \end{minipage}%
        \begin{minipage}{0.24\linewidth}
        \centering
        \includegraphics[width=\linewidth]{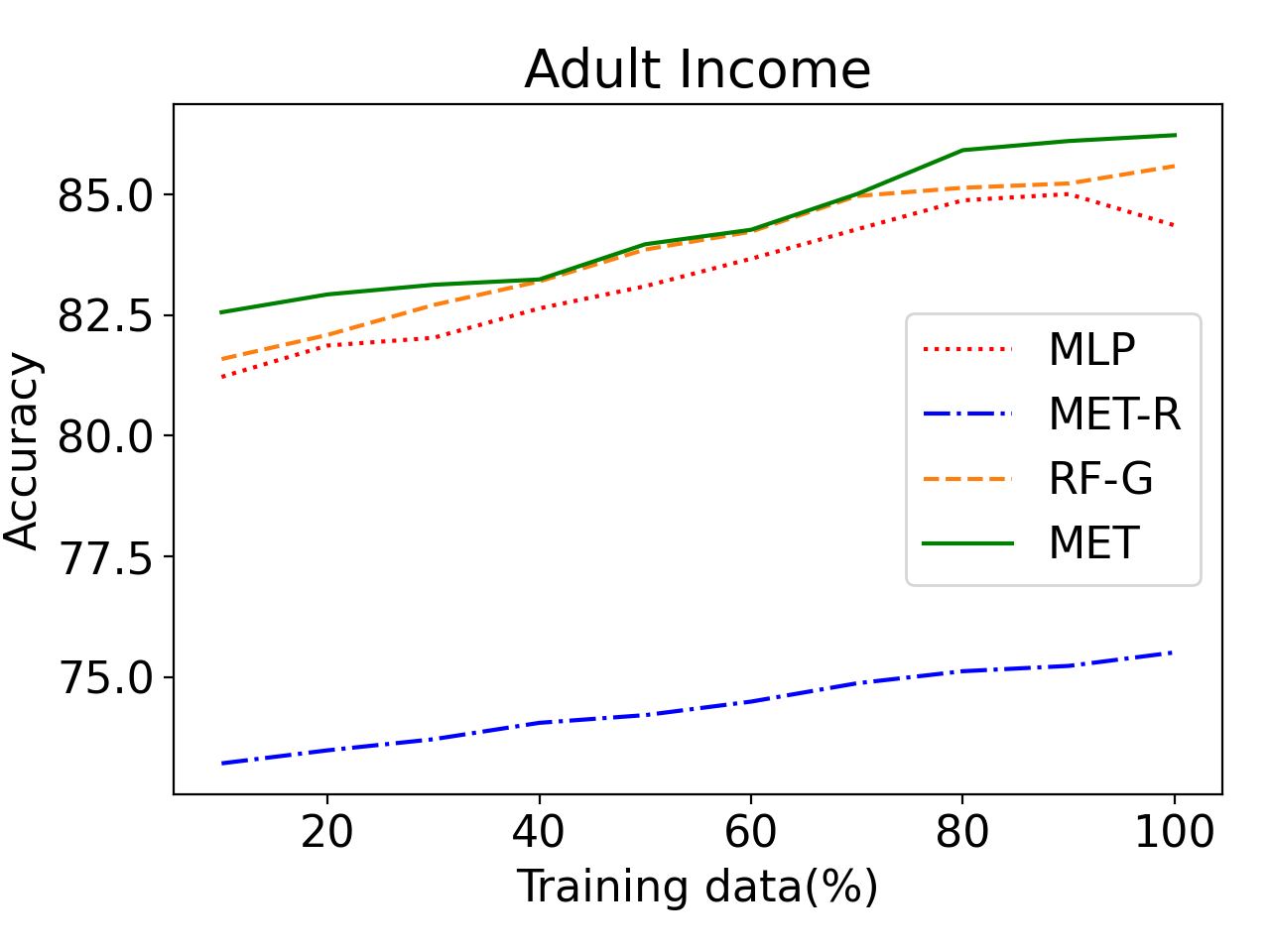}
    \end{minipage}\vspace*{-5pt}
\caption{We compare the performance (downstream classification accuracy) of \method with various baselines as the fraction of labelled data used for training the downstream classifier is varied. Observe that \method consistently outperforms the baselines even when a smaller fraction of labelled data is used for downstream classifier training.}
\label{fig:trn_pct}
\end{figure}

\textbf{Accuracy with a fraction of labeled training data}:
Next we compare our proposed algorithm \method against the baselines when only a fraction of labelled data is used for the supervised training. Specifically, we vary the fraction of labeled data used for training the downstream classifier from $20\%$ to $100\%$ and compare the obtained downstream classification accuracies with the baselines. \autoref{fig:trn_pct} shows the variation of accuracy with the fraction of labeled data for \method, comparing against the baselines like gaussian random featurization (RF-G), learning MLP directly over raw features and a random encoder (MET-R). We observe that \method outperforms the baselines for all the choices of fraction of labeled data used for supervised learning.

\textbf{Concatenated vs averaged embeddings}: Next, we try two approaches for the task of getting good tabular representations from co-ordinate level representations: 
\begin{itemize}[leftmargin=*]
    \item \textbf{Concatenation} : Concatenating co-ordinate level representations learnt by \method to represent tabular level representations.
    
    \item \textbf{Averaging} : Taking an average of representations learnt by \method over all co-ordinates to represent tabular level representations.
\end{itemize}
We try above mentioned approaches for \fmnist and \covtype. For both datasets, concatenation significantly outperforms averaging. For  \covtype, concatenation and averaging obtain $74.11\%$ and $61.87\%$ accuracies respectively and for \fmnist, they obtain $90.94\%$ and $88.64\%$ respectively.

\textbf{Effect of adversarial reconstruction}: Next, we analyze the effect of using gaussian noise along with masking for reconstruction based SSL on tabular datasets. In \autoref{tab:classification}, observe that \method always outperforms or matches \methodstd (without adversarial noise), and in some cases like \covtype gives upto $2.5\%$ improvement compared to the non-adversarial counterpart.

\begin{table}
  \centering
\begin{tabular}{|l|lrrrrr|}
\hline
\textbf{Type}                            & \textbf{Methods} & \multicolumn{1}{|l}{FMNIST} & \multicolumn{1}{|l}{CIFAR10} & \multicolumn{1}{|l}{MNIST} & \multicolumn{1}{|l}{CovType} & \multicolumn{1}{|l|}{Income} \\ \hline \hline
\multirow{3}{*}{Supervised Baseline}     & MLP     & 87.62\%                    & 16.50\%                     & 96.95\%                   & 65.47\%                     & 84.36\%                    \\ \cline{2-7} 
                                         & RF             & 87.19\%	& 36.75\%	& 97.62\%	& 65.00\%	& 84.60\%                    \\ \cline{2-7} 
                                         & RF-G             & 89.84\%                    & 29.32\%                     & 97.65\%                   & 71.57\%                     & 85.59\%                    \\ \cline{2-7} 
                                         & MET-R            & 88.84\%                    & 28.94\%                     & 97.44\%                   & 69.68\%                     & 75.51\%                    \\ \hline \hline
\multirow{3}{*}{Self-Supervised Methods} & VIME             & 80.36\%                    & 34.00\%                        & 95.77\%                   & 62.80\%                     & 86.00\%                       \\ \cline{2-7} 
                                         & DACL+            & 81.40\%                    & 39.70\%                     & 91.40\%                   & 64.23\%                     & 84.47\%                    \\ \cline{2-7} 
                                         & SubTab           & 87.59\%                    & 39.34\%                     & 98.31\%                   & 42.36\%                     & 84.43\%                    \\ \hline \hline
\multirow{2}{*}{Our Method}              & MET-S            & 90.94\%                    & \textbf{48.00\%}            & 99.01\%                   & 74.11\%                     & \textbf{86.23\%}           \\ \cline{2-7} 
                                         & MET              & \textbf{91.36\%}           & 47.82\%                     & \textbf{99.19\%}          & \textbf{76.71\%}            & 86.25\%                    \\ \hline
\end{tabular}\vspace*{5pt}
\caption{Downstream classification accuracy on five common tabular datasets, comparing \method against various baselines. \method uses adversarial training + masking for reconstruction based self-supervised-learning whereas \methodstd is an ablation where only masking is used. \method outperforms the baselines across the all the datasets, giving an average gain of $3.2\%$ compared to the best competitive baseline for each dataset. Results with the standard deviations are in the appendix. Although the maximum deviation is not significant enough.}
\label{tab:classification}
\end{table}

\subsection{Ablation Studies}
In this section, we perform an ablation study to understand the effect of various aspects of \method such as masking ratio, depth of the MLP over learnt representations and  how the masked tokens are used. 
\paragraph{Masking Ratio} \label{sec:masking_ratio}
As discussed in Section \ref{sec:method}, \method relies on the use of masked reconstruction along with adversarial noise for learning well separated representations. We first analyze the performance of \method as the masking ratio (fraction of the input tokens masked) is varied, keeping all the other hyper-parameters fixed. Our observations yield that in general, a high masking ratio ($50\%-70\%$) seems to give the best downstream accuracy and hence the best learnt representations. For example, on \fmnist, \mnist and \adult a masking ratio of $70\%$ works the best whereas on \covtype a masking ratio of $50\%$ gives the best results. \autoref{fig:mask_pct} in Appendix \ref{app:masking} shows detailed plots of change in accuracy as masking ratio is varied across four datasets.

\paragraph{Depth of Finetuning MLP Head}
Next we analyze how sensitive the downstream task performance is to variation of the depth of the MLP over the learnt representations. We fix the learnt representations for the \fmnist dataset, and vary the number of hidden layers in the MLP. \autoref{tab:mlp_depth} shows the variation of accuracy with the hidden layers. We find that for most of the datasets, a depth of two to three hidden layers works the best. Although, on the same note, we mention that the downstream accuracy is not much sensitive to the choice of number of hidden layers. For example, on \fmnist the best accuracy is $90.86\%$ with 2 hidden layers whereas with 5 hidden layers, the accuracy is $90.62\%$.

\begin{table}[h]
  \centering
    \begin{tabular}{|c|c|c|c|c|c|c|}
    \hline
    \textbf{Num hidden layers} & 0 & 1&2&3&4&5\\\hline 
    \textbf{Accuracy} & 87.16\%& 90.59\%& 90.86\%& 90.82\%& 90.65\%& 90.62\%  \\\hline
    \end{tabular}\vspace*{5pt}
  \caption{Downstream task accuracy with varying number of hidden layers in the downstream MLP classifier with the learnt representation being fixed. We observe that for most datasets $2$ or $3$ hidden layers achieve the best performance, with relatively low sensitivity to the choice of number of hidden layers.}
  \label{tab:mlp_depth}
\end{table}

\vspace{-1em}
\paragraph{Mask Token}
Recall that \method uses a learnable mask token, which is passed to the decoder directly. Note that the mask token is kept same for all the masked co-ordinates (although positional encoding would be different) in \method. Now, consider the following two variants of \method: 
\begin{itemize}[leftmargin=*]
    \item MET-separate-token : a separate mask token is learned for each  co-ordinate, which is passed  directly to the decoder.
    
    \item MET-encoder : the masked token is passed through the encoder before being fed into the decoder. 
    
\end{itemize}

We compare \method against both the above version of \method on the two datasets \fmnist and \covtype. Our analysis highlights some interesting aspects about \method. 

First, we compare \method against MET-separate-token (as described above). We observe that on \fmnist both \method ($90.94\%$) and MET-separate-token ($90.90\%$) perform almost similarly, although interestingly, MET-separate-token performs more than $1\%$ better than \method on \covtype. We hypothesize the reason for this being the fact that on \fmnist, all the co-ordinates are pixels which belong to the same domain whereas on \covtype features come from different domains. Hence using a separate mask token for each feature can potentially give improvements as observed in our experiments. However, on the same note, the memory requirement for having a different mask token for all features is very large and hence, we use the same mask token for all the features in \method.

Next, we compare \method to MET-encoder where we pass the learnt mask token through the encoder too; see Table~\ref{tab:mask_set}. We observe that passing mask token through the encoder does not lead to any performance gains on \fmnist and leads to performance degradation on \covtype. Further, passing the masked token through the encoder increases the compute requirements. Hence, similar to the observations in \cite{mae}, we chose to pass the mask token directly to the decoder. 

\begin{table}[]
    \centering
        \begin{tabular}{|l|c|c|c|}
    \hline
            & \multicolumn{1}{l}{\textbf{MET}} & \multicolumn{1}{|l}{\textbf{MET-encoder}} & \multicolumn{1}{|l|}{\textbf{MET-seperate-token}} \\ \hline \hline
    \textbf{FMNIST}  & 90.94\%                          & 90.92\%                                  & 90.90\%                                         \\ \hline
    \textbf{CovType} & 74.12\%                          & 73.70\%                                  & 75.40\%                                         \\ \hline
        \end{tabular}\vspace*{5pt}
    \caption{Comparing \method with it's different versions : (1) MET-encoder where we pass the mask token through the encoder before giving as an input to the decoder. (2) MET-separate-token where every feature has a separate learnable mask token. We observe that for dataset like \covtype where every feature belongs to different domains, learning separate token can potentially give further performance gains at the cost of memory.}
    \label{tab:mask_set}
\end{table}


\section{Conclusion and Limitations}\label{sec:limitation}
In this paper, we proposed a \emph{purely reconstruction based} SSL algorithm, \method, for representation learning on tabular datasets. The two key ideas in \method are (i) use a concatenation of representations for all features instead of averaging, and (ii) use adversarial reconstruction loss in addition to the standard loss. Through experiments on five tabular datasets, we showed  that \method achieves a new SOTA result for downstream classification on these datasets, improving over previous contrastive based approaches by $3.2\%$ on average. 

While reconstruction based SSL has been shown to learn powerful representations across various domains such as text~\cite{bert}, vision~\cite{mae} as well as tabular (this paper), a thorough understanding of \emph{why} reconstruction loss promotes linearly separable representations is  missing in the literature. In this paper, we took  a step forward by proposing a simple dataset, that can act as a test bed for answering this question. We showed empirically that while this dataset is not linearly separable in the input space, it became linearly separable using \method representations. We believe that a mathematical proof of this phenomenon could shed light on \emph{why} and \emph{how} reconstruction based approaches learn useful representations. 
We also demonstrated that using a concatenation of representations of all features/coordinates gives substantially better results than pooling of all token level representations as done in vision~\cite{mae} and text~\cite{bert}. This is intuitive as the average representation of all tokens is not specifically trained to be useful for the downstream task. At the same time, concatenation significantly increases the representation dimension as well as the complexity of the downstream finetuning model. Hence, it can  exacerbate the risk of overfitting. For tabular-SSL, our results ruled out this case. However, a thorough investigation of this aspect is also an interesting direction for future work. 

\bibliography{ref}
\bibliographystyle{plain}
\newpage
\appendix
\section{Appendix}

\subsection{Masking Ratio} \label{app:masking}
\begin{figure}[h]
    \begin{center}
    \includegraphics[width=\linewidth]{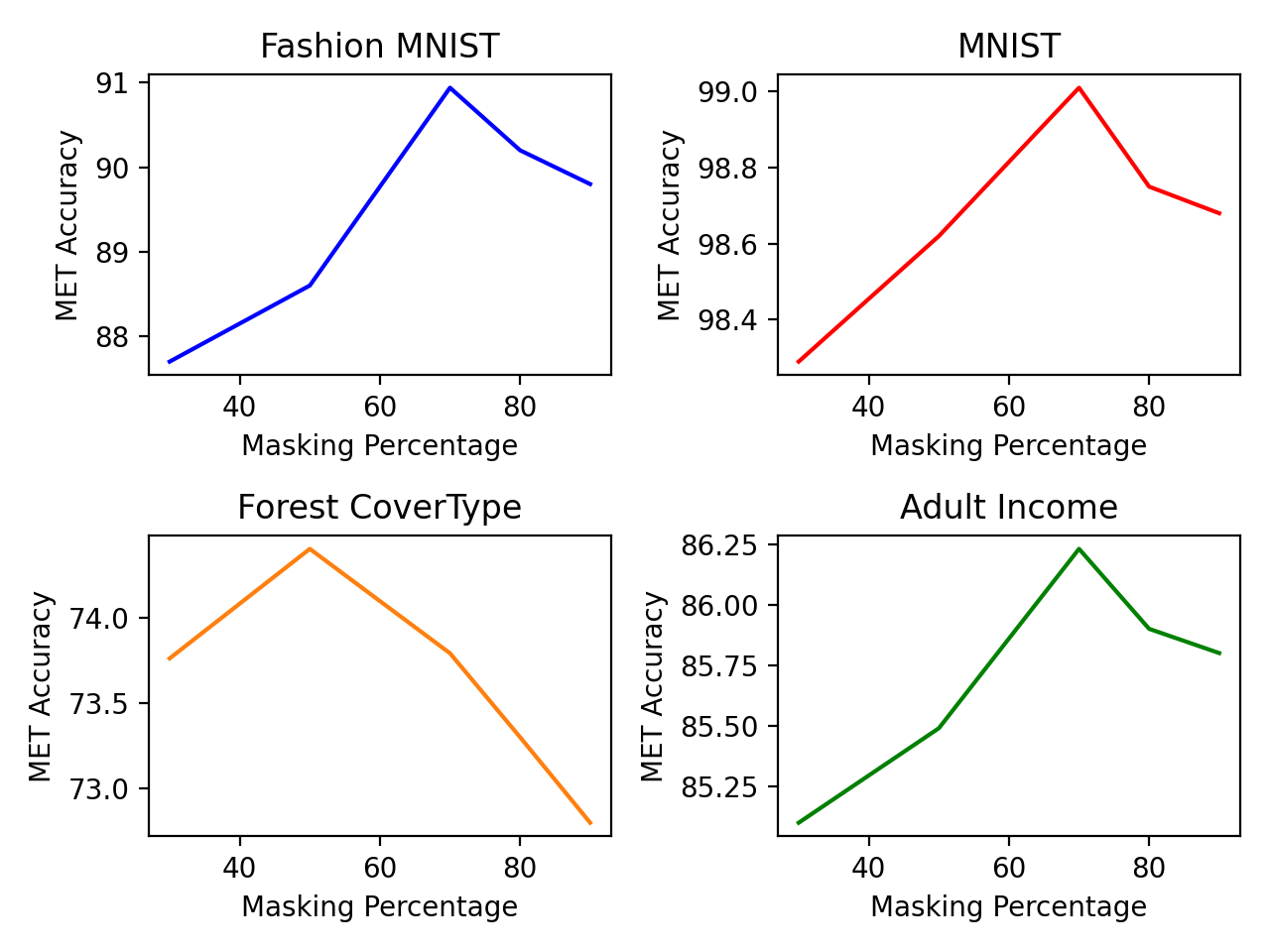}
      \caption{We study the variation in downstream accuracy as the masking ratio is varied in \method, for four tabular classification datasets. We observe that a high masking ratio ($50\%$-$70\%$) generally works the best.}
    \label{fig:mask_pct}
\end{center}
\end{figure}

We discussed the variation in performance of \method as we vary the fraction of input tokens masked in Section~\ref{sec:masking_ratio}. Our general observation is that \method learns the best representations (in terms of downstream classification accuracy) when the masking ratio is high ($50\%$-$70\%$). \autoref{fig:mask_pct} shows the variation of downstream classification accuracy as the masking ratio is varied in $\{30, 50, 70, 80, 90\}$.

\subsection{Hyper-Parameters}
In this section, we share the exact hyper-parameters of \method for replicating the results on all the five tabular datasets. Note that Encoder Depth refers to the number of transformer layers in the encoder stack and Decoder Depth refers to the number of transformer layers in the decoder stack. Adversarial Learning Rate ($lr\_adv$) refers to the learning rate used for gradient ascent in adversarial loop and Learning Rate($lr$) refers to the learning rate for gradient descent on reconstruction loss. We perform a grid-search for Embedding dimension($e$) and Feed-forward dimension($fw$) in $\{64,100,128\}$, Number of Heads in the transformer architecture in $\{1,2,3\}$, encoder and decoder depth in $\{1,3,6\}$, Learning Rate($lr$) in $\{1e^{-1},1e^{-2},1e^{-3},1e^{-4},1e^{-5}\}$, masking percentage($m$) in $\{30,50,70,80,90\}$, adversarial steps($adv\_steps$) in $1,2,4$, radius of L2-norm ball($\epsilon$) in $\{2,6,10,12,14\}$ and learning rate of gradient ascent step($lr\_adv$) in $\{0.1,0.01\}$. The optimal set of hyper-parameters for various datasets obtained are mentioned in Tables\ref{tab:hyp_part1} and \ref{tab:hyp_part2}. Note that all the ablation studies are conducted using these parameters and \methodstd.

\begin{table}[h]
    \centering
        \begin{tabular}{|c|c|c|c|c|c|}
        \hline
                               & \multicolumn{1}{c|}{\begin{tabular}[c]{@{}c@{}}Embedding \\ Dimension($e$)\end{tabular}} & \multicolumn{1}{c|}{\begin{tabular}[c]{@{}c@{}}Feed-forward \\ Dimension($fw$)\end{tabular}} & \multicolumn{1}{c|}{\begin{tabular}[c]{@{}c@{}}Number of \\ Heads\end{tabular}} & \multicolumn{1}{c|}{\begin{tabular}[c]{@{}c@{}}Encoder \\ Depth\end{tabular}} & \multicolumn{1}{c|}{\begin{tabular}[c]{@{}c@{}}Decoder \\ Depth\end{tabular}} \\ \hline \hline
        Fashion MNIST & 64                                                   & 64                                                 & 1                                             & 6                                           & 1                                           \\ \hline
        CIFAR10       & 100                                                  & 64                                                 & 2                                             & 3                                           & 3                                           \\ \hline
        MNIST         & 64                                                   & 64                                                 & 1                                             & 6                                           & 1                                           \\ \hline
        CovType       & 100                                                  & 64                                                 & 1                                             & 1                                           & 1                                           \\ \hline
        Adult Income  & 64                                                   & 64                                                 & 1                                             & 3                                           & 6                                           \\ \hline
        \end{tabular}
    \vspace{1ex}
    \caption{We share the exact hyper-parameters for replicating the results with \method.}
    \label{tab:hyp_part1}
\end{table}

\begin{table}[h]
    \centering
        \begin{tabular}{|c|c|c|c|c|c|}
        \hline
                      & \multicolumn{1}{c|}{$lr(\eta_{1})$} & \multicolumn{1}{c|}{\begin{tabular}[c]{@{}c@{}}Masking \\ Percentage($m$)\end{tabular}} & \multicolumn{1}{c|}{\begin{tabular}[c]{@{}c@{}}Adversarial \\ Steps($adv\_steps$)\end{tabular}} & \multicolumn{1}{c|}{\begin{tabular}[c]{@{}c@{}}L2 Norm Ball \\ Radius($\epsilon$)\end{tabular}} & \multicolumn{1}{c|}{\begin{tabular}[c]{@{}c@{}}Adversarial $lr$ \\($\eta_{2}$)\end{tabular}} \\ \hline \hline
        Fashion MNIST & $1e^{-5}$                               & 70                                  & 2                                                  & 2                                                                 & $1e^{-2}$                                                \\ \hline
        CIFAR10       & $1e^{-4}$                               & 70                                  & 3                                                  & 14                                           & $1e^{-2}$                                                \\ \hline
        MNIST         & $1e^{-4}$                               & 70                                  & 2                                                  & 12                                                                & $1e^{-2}$                                                \\ \hline
        CovType       & $1e^{-4}$                               & 50                                  & 5                                                  & 4                                                                 & $1e^{-1}$                                                \\ \hline
        Adult Income  & $1e^{-3}$                               & 80                                  & 1                                                  & 6                                                                 & $1e^{-1}$                                                \\ \hline
        \end{tabular}
    \vspace{1ex}
    \caption{We share the exact hyper-parameters for replicating the results with \method.}
    \label{tab:hyp_part2}
\end{table}

\subsection{Mean and Standard Deviation for Downstream Classification Results}
In Section~\ref{sec:down_acc}, we compared the downstream classification accuracy for various baselines and \method over five tabular datasets (\autoref{tab:classification}). For completeness, we add the results with mean and standard deviations, computed over five independent runs of each algorithm in \autoref{tab:classification_stddev}.

\begin{table}[h]
\footnotesize
    \centering
\begin{tabular}{|l|l|c|c|c|c|c|}
\hline
\textbf{Type}                            & \textbf{Methods} & \multicolumn{1}{|l}{FMNIST} & \multicolumn{1}{|l}{CIFAR10} & \multicolumn{1}{|l}{MNIST} & \multicolumn{1}{|l}{CovType} & \multicolumn{1}{|l|}{Income} \\ \hline \hline
\multirow{3}{*}{Supervised Baseline}     & MLP     & 87.57 $\pm$ 0.13                    & 16.47 $\pm$ 0.23                     & 96.98 $\pm$ 0.1                   & 65.45 $\pm$ 0.09                     & 84.35 $\pm$ 0.11                    \\ \cline{2-7} 
                                         & RF             & 87.19 $\pm$ 0.09	& 36.75 $\pm$ 0.17 &	97.62 $\pm$ 0.18	& 64.94 $\pm$ 0.12 &	84.60 $\pm$ 0.2                    \\ \cline{2-7} 
                                         & RF-G             & 89.84 $\pm$ 0.08 &	29.28 $\pm$ 0.16 &	97.63 $\pm$ 0.03 &	71.53 $\pm$ 0.06	& 85.57 $\pm$ 0.13                    \\ \cline{2-7}                                          & MET-R            & 88.81 $\pm$ 0.12 &	28.97 $\pm$ 0.08 &	97.43 $\pm$ 0.02 &	69.68 $\pm$ 0.07	& 75.50 $\pm$ 0.04                    \\ \hline \hline
\multirow{3}{*}{Self-Supervised Methods} & VIME             & 80.36 $\pm$ 0.02 &	34.00 $\pm$ 0.5 &	95.74 $\pm$ 0.03 &	62.78 $\pm$ 0.02	 & 85.99 $\pm$ 0.04                       \\ \cline{2-7} 
                                         & DACL+            & 81.38 $\pm$ 0.03 &	39.7 $\pm$ 0.06 &	91.35 $\pm$ 0.075 &	64.17 $\pm$ 0.12 &	84.46 $\pm$ 0.03                   \\ \cline{2-7} 
                                         & SubTab           & 87.58 $\pm$ 0.03 &	39.32 $\pm$ 0.04 &	98.31 $\pm$ 0.06 &	42.36 $\pm$ 0.03 &	84.41 $\pm$ 0.06                    \\ \hline \hline
\multirow{2}{*}{Our Method}             & MET-S            & 90.90 $\pm$ 0.06 &	\textbf{47.96  $\pm$  0.1} &	98.98 $\pm$ 0.05	& 74.13 $\pm$ 0.04	& \textbf{86.17  $\pm$  0.08}           \\ \cline{2-7} 
                                         & MET              & \textbf{91.32} $\pm$ 0.08	& \textbf{47.92  $\pm$  0.13} &	\textbf{99.17 $\pm$ 0.04} &	\textbf{76.68  $\pm$  0.12} & 	\textbf{86.21 $\pm$ 0.05}                    \\ \hline
\end{tabular}
\vspace{1ex}
\caption{Downstream classification accuracies over five commonly used tabular benchmarks. \method uses adversarial training + masking for reconstruction based self-supervised learning, whereas MET-S is an ablation where only masking is used. We observe that \method outperforms the baselines over all the datasets, giving an average improvement of $3.2\%$ compared to the most competitive baseline for each of the dataset.}
    \label{tab:classification_stddev}
\end{table}
\end{document}